\pdfoutput=1
\documentclass[lettersize,journal]{IEEEtran}
\usepackage[T1]{fontenc}
\usepackage[utf8]{inputenc}
\usepackage{amsmath,amssymb,amsfonts}
\usepackage{mathptmx}
\DeclareMathAlphabet{\mathcal}{OMS}{cmsy}{m}{n}
\usepackage{array}
\usepackage{textcomp}
\usepackage{stfloats}
\usepackage{url}
\usepackage{graphicx}
\usepackage{cite}
\usepackage{booktabs}
\usepackage{tabularx}
\usepackage{microtype}
\usepackage[hidelinks]{hyperref}
\hypersetup{pdftitle={MouldIt: Data Efficiency of Multi-Task Temporal Representation Learning in Injection Moulding},pdfauthor={Abdullah Ayad}}
\newcommand{\doi}[1]{doi:~\href{https://doi.org/#1}{\nolinkurl{#1}}}


\AtBeginDocument{\lineskiplimit=-2.5pt\relax}
\makeatletter
\g@addto@macro\normalsize{%
  \setlength\baselineskip{12pt plus 0.35pt}%
  \setlength\abovedisplayskip{6pt plus 1pt minus 1pt}%
  \setlength\belowdisplayskip{6pt plus 1pt minus 1pt}%
  \setlength\abovedisplayshortskip{3pt plus 1pt}%
  \setlength\belowdisplayshortskip{4pt plus 1pt minus 1pt}}
\makeatother
\begin{document}

\title{MouldIt: Data Efficiency of Multi-Task Temporal Representation Learning in Injection Moulding}

\author{Abdullah~Ayad
\thanks{A. Ayad (e-mail: \mbox{\href{mailto:abdallahayman772@gmail.com}{abdallahayman772@gmail.com}}) is an independent researcher. Most of this work was done while with the Department of CSAI, University of Technology Nuremberg, 90461 Nuremberg, Germany.}
\thanks{This work has been submitted to the IEEE for possible publication. Copyright may be transferred without notice, after which this version may no longer be accessible.\looseness=-1}}

\markboth{Preprint submitted to IEEE Transactions on Automation Science and Engineering}%
{Ayad: MouldIt: Data Efficiency of Multi-Task Temporal Representation Learning in Injection Moulding}

\maketitle

\begin{abstract}
Smart industrial solutions that can address multiple tasks require models that understand the underlying processes. We examine representation learning for modeling the injection moulding process as an example of such solutions. We consider its data efficiency and multi-modal benefit compared to classical machine learning alternatives. To this end, we synthesize a process-informed dataset based on a real reference cycle of an Erlanger Träger part. We then introduce MouldIt, a temporal convolution and attention based variational autoencoder that reconstructs in-mould signals, while classifying parts as acceptable or as one of our synthetic failure modes. We evaluate MouldIt on two tasks, known-class classification and unknown-class detection, across nine dataset sizes in single-modal and multi-modal settings. We compare it with an accept-all baseline and the classical methods k-nearest neighbors, logistic regression, random forests, and principal component analysis, which we adapt to support both tasks. MouldIt outperforms classical methods across both tasks from 4000 cycles in the single-modal setting and from 2000 cycles in the multi-modal setting. It achieves part sorting losses of 6.0\,\textpm\,0.7\,\% and 1.0\,\textpm\,0.3\,\% in the single-modal and multi-modal settings respectively, ranking first across all baselines and a combination of principal component analysis and random forests.
\end{abstract}

{\renewcommand{\abstractname}{Note to Practitioners}
\begin{abstract}
Sorting injection moulded parts automatically requires a system that recognizes common defects and also flags defects it has never seen. Defect data are expensive to collect, so engineers need to know how much data such a system requires and whether additional sensors are worth installing. Better sorting raises quality by shipping fewer faulty parts and reduces waste by rejecting fewer good ones, and one model that handles both recognition and flagging avoids maintaining two. We studied these questions on synthetic cycles generated from one recorded cycle of a real part. With only a cavity pressure sensor, random forests with a separate check for unknown defects sort parts best when data are scarce, and our neural network overtakes them only from about 4000 cycles. With a temperature sensor and two ultrasound sensors added, the network sorts best at every dataset size we tested, although the random forests match it at 16000 cycles. These results can guide a first choice of sensors and methods, but do not replace testing on a real line: all cycles derive from one part and one material, the synthetic defect sizes are modeling choices, and about 5\% false alarms is high for production. Before deployment, the system should be validated on real defects, and its alarm threshold set by weighing the cost of shipping a faulty part against that of rejecting a good one. The learned representations of the process could also be used for control, not only for sorting.
\end{abstract}}

\begin{IEEEkeywords}
Data efficiency, fault detection, injection moulding, multi-modal sensing, part sorting, representation learning, synthetic data, variational autoencoders
\end{IEEEkeywords}

\section{Introduction}

\label{sec:introduction}
\IEEEPARstart{I}{ndustrial} production has traditionally relied on custom pre-programmed solutions that operate in known and optimized environments with reliability, productivity, and cost reduction as their objectives. Meanwhile, in other domains like language, general-purpose solutions have offered us a different problem-solving perspective. Instead of targeting a single objective, language models aim to understand their data in a way that allows them to address multiple downstream tasks. Inspired by this progress, we expect similar benefits from learning representations of industrial processes. 

Learning such representations has to account for the nature of industrial process data, which are generally temporal and usually tabular. Temporal neural network architectures with suitable inductive biases can offer more than the prediction strengths of classical methods like random forests or the encoding strengths of methods like PCA.

\begin{figure}[!t]
    \centering
    \includegraphics[width=\columnwidth]
        {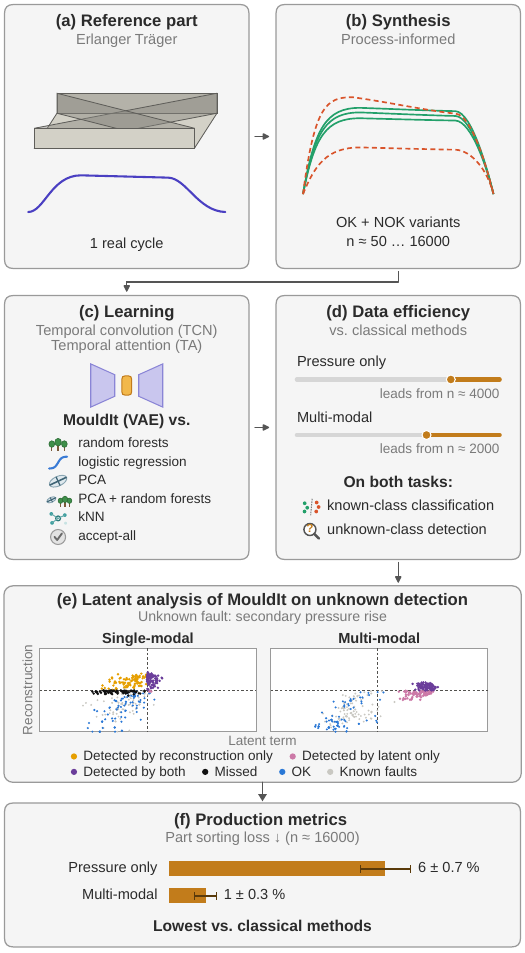}
    \caption{Overview of this work. (a) We start from one real cycle of an Erlanger Träger part. (b) From it, we synthesize OK cycles and process-informed NOK variants, in datasets of 50 to 16000 cycles. (c) We train MouldIt, a temporal VAE with convolution and attention layers, and compare it with random forests, logistic regression, PCA, PCA with random forests, kNN, and an accept-all reference. (d) MouldIt leads on both known-class classification and unknown-class detection from 4000 cycles with pressure only, and from 2000 cycles with multi-modal input. (e) A latent analysis shows whether each unknown cycle is detected by the latent term, the reconstruction term, or both, here for secondary pressure rise. (f) At 16000 cycles, its part sorting loss is $6 \pm 0.7\%$ with pressure only and $1 \pm 0.3\%$ with multi-modal input.}
    \label{fig:teaser}
\end{figure}

Injection moulding is a suitable candidate for evaluating such models, as it is a critical process for most industries, and it typically employs multi-modal data. The process is characterized by static machine setting parameters (MSPs) and dynamic process parameters (PPs), typically recorded on the producing machines. These process parameters are commonly used for tasks such as monitoring for anomaly detection, process optimization, and process control. For fine-grained dynamic information, in-mould sensors are usually installed, such as pressure, temperature, and ultrasound sensors. Ultrasound, for example, can track gap development between the part and the cavity wall, which pressure cannot monitor \cite{kariminejad2021USinIM}. The process outputs usually have a number of quality metrics, like weight, size, or specific part deformation labels.

A natural application in this process is automated part sorting, where a decision can go wrong in either direction: good parts are rejected, or faulty parts are shipped. Sorting builds on two tasks, classifying known faults and detecting unknown ones, since a classifier trained on known faults alone can ship a new fault as a good part. A single model that performs both tasks would also avoid maintaining separate models. Automated sorting thus addresses three concerns of production: quality, by keeping faulty parts from shipping; efficiency, by not rejecting good ones; and reliability, by flagging unseen faults.

Naturally, the wide adoption of such models in industry faces barriers. These models are unpredictable and not well understood; therefore, works on explainability, on necessary conditions for those models to excel, and on production metrics may facilitate their adoption.

Data efficiency is among these necessary conditions, since representation learning typically requires large amounts of data to learn the underlying process. This raises four questions: how much data is needed for model performance to saturate; whether incorporating additional modalities can reduce this requirement; whether, at such a saturation point, representation-learning models improve over classical alternatives on production-relevant metrics; and whether their learned representations generalize to unknown failure modes.

Answering these questions requires data that cover many failure modes. However, since representation learning is not yet widely used in this industry, foundational injection moulding datasets are not yet available. Additionally, access to industry-grade simulation software is often proprietary. Synthetic data is thus a practical way to explain and benchmark such models.

We therefore address them on synthetic injection moulding data (Fig.~\ref{fig:teaser}), with the following contributions:
\begin{itemize}
    \item an injection moulding data synthesis method requiring a single real sample, inspired by the success of \cite{hollmann2023tabpf} with synthetic training.
    \item a novel multi-objective temporal representation learning model for injection moulding in-mould data (pressure, temperature, and ultrasound).
    \item an analysis of the effect of dataset size and input modalities on model performance across known-class classification and unknown-class detection.
    \item a latent analysis of how MouldIt detects unknown faults.
    \item an evaluation of part sorting losses across the same dataset sizes and input modalities.
\end{itemize}

\section{Related Works}
\label{sec:related}
Our work focuses on representation learning for industrial applications, specifically injection moulding. We therefore review each topic first across industrial processes, then narrow down to injection moulding where such work exists.

\subsection{Synthetic Data and Injection Moulding Datasets}
Synthetic data generation has been explored in manufacturing to address the scarcity of defective samples, including methods that generate visual defects for industrial inspection \cite{Mou2025synth4seg,li2024dlsgan,tai2025diffusion}.

In injection moulding (IM), public datasets are scarce and heterogeneous. Some provide cycle-level scalar features \cite{polenta22MLinj} or machine-side time series such as injection pressure and flow curves \cite{Bogedalefinegrainedseries,west_2025_dataset_fine_grained}. The dataset \cite{lambers_2022_fk2gn-x8e76} additionally contains full-cycle cavity pressure (in-mould), with manual OK/NOK and defect-type labels. Further in-mould pressure data are either limited to a few labeled curves \cite{ahmed_shaikh_2024_inmould_dataset} or simulated \cite{michiels2022simualationinmouldqualpred}. Synthetic generation of IM data from causal models has also been explored \cite{HENNEBOLD2025syntheticdatacausal}, though only for cycle-level scalar features rather than full-cycle pressure curves. We consider an Erlanger Träger part, for which a public dataset provides one cycle of in-mould pressure, temperature and ultrasound data \cite{kini2026forinfpro}. We therefore synthesize a larger dataset from this cycle.

\subsection{Representation Learning for Industrial Processes}
Representation learning has gained traction for modeling industrial processes. In fault diagnosis, \cite{T-ASE_multi_modal_fusion_repl_industry} compares single-modal and multi-modal inputs and explores how to fuse them. Autoencoders have been used to detect anomalies in semiconductor manufacturing \cite{maggipinto2022autoencoder} and, through their latent space, faults in air handling units \cite{tra2026latent}, while multi-modal fusion has supported soft sensing in sintering \cite{zhang2026multimodal}.

In injection moulding, classical machine learning models, specifically random forests, have been effective on tabular data \cite{polenta22MLinj}. More recently, representation learning, which learns latent process representations alongside task-specific predictions, has been used for process monitoring and anomaly detection \cite{yaninj, Vilianomalyvae, kim21faultdetecrep}, quality prediction \cite{MUAZ2023238}, and process control \cite{yan2025learningactionableworldmodels}. We investigate it through two new lenses: data efficiency and multi-modal benefit. Unlike \cite{yaninj}, which uses a single modality, we design a novel multi-objective VAE that learns from multi-modal synthetic data, and we study both lenses on two tasks: classification and unknown-class detection.

\subsection{Data Efficiency of Industrial Models}
Labeled IM data require production trials and part inspection, which makes the amount of training data a central constraint. Many works therefore reduce the data a model needs. In manufacturing, a model can learn from the quality estimates of physics-based models when measured labels are costly \cite{alenezi2025physics}. Transfer learning has also been used to generate scarce fault data \cite{jiang2025transfer}. In IM, transfer learning of supervised models on scalar process parameters is a common approach \cite{TERCAN2018transferlearningscalar, Lockner2021transferbetmachinesandparts, LocknerbetweenmaterialsTL}.

A separate question is how much data a model needs in the first place, which shows in how its performance scales with the training-set size. For cavity pressure data, a decision tree has reached over 90\% accuracy with 1–3 training samples per class \cite{mlcavitypressuredataefficiency}. For learned representations, this scaling has not been the focus of prior IM studies, neither for classification accuracy nor for unknown-class detection. We measure this scaling for MouldIt and classical methods.

\subsection{Unknown Fault Detection in Industrial Processes}
Production can encounter faults that are missing from the training data. \cite{zuo2025openset} address open-set fault detection with a contrastive conditional adversarial autoencoder, and \cite{liu2026zeroshot} diagnose unseen faults from fault attributes. \cite{cai2025unified} unify fault detection and classification, which are conventionally treated as separate tasks, in a framework that handles new faults without full retraining. We consider these tasks in injection moulding, with a single model that classifies known faults and detects unknown ones.\looseness=-1

\subsection{Explainability of Industrial Models}
Interpretable fault diagnosis has built on process knowledge \cite{he2025mqkin} and on the dynamic dependencies between process variables \cite{ma2026interpretable}. In IM, \cite{GimCavityPressAnalysis} interpret a neural network to analyze how features of the cavity pressure profile affect part weight, \cite{Vilianomalyvae} interpret the root causes of detected anomalies, and \cite{yaninj} derive interpretable indices of the deviations caused by machine parameters. We analyze the latent space to explain how MouldIt detects unknown faults.

\section{Methods}

\subsection{Synthetic Data Generation}
\label{subsec:synthetic-data-generation}
We generate synthetic single-modal and multi-modal datasets and label each sample as OK or as one of several process-informed NOK variants.
We first preprocess the reference cycle by aligning its modalities on a common time grid (Section~\ref{methods:temporal_alignment}).
We then locate phase markers (fill start, pressure peak, holding end) in the normalized cavity pressure channel, as well as the temperature peak and the ultrasonic minima (Section~\ref{methods:phase extraction}); most failure modes act relative to these points.
Next, we initialize each sample as an OK copy of the reference cycle and obtain each NOK variant by applying one of the failure modes of the respective setting (Sections~\ref{methods:single} and~\ref{methods:multi}).
Finally, we apply OK disturbances, i.e., benign process variations that leave the label unchanged, to all samples.

\subsubsection{Input Modalities}
\begin{figure}[!t]
    \centering
    \includegraphics[width=\columnwidth]
        {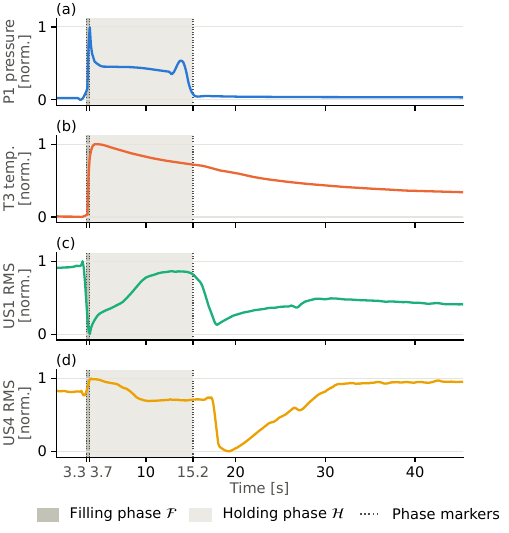}
    \caption{Real Erlanger Träger reference cycle from which all cycles are synthesized, on the common time grid and with each channel min-max normalized. The single-modal input is (a) the cavity pressure of Kistler sensor P1, and the multi-modal input adds (b) the cavity temperature of Kistler sensor T3 and (c), (d) the RMS amplitudes of ultrasound sensors US1 and US4. Dotted lines mark the fill start, pressure peak, and holding end at 3.3, 3.7, and 15.2\,s, and shaded regions the filling and holding phases. The temperature peaks at 4.4\,s, US1 has its minimum at the pressure peak, and US4 has its minimum in the cooling phase at 19.3\,s.}
    \label{fig:modalities}
\end{figure}

Our reference cycle is the first cycle of a public dataset of an Erlanger Träger part \cite{kini2026forinfpro}. In the single-modal setting, we learn from Kistler in-mould pressure sensors (Fig.~\ref{fig:modalities}(a)), as in \cite{yaninj}. In the multi-modal setting, we use temperature, pressure, and two ultrasound sensors (Fig.~\ref{fig:modalities}).

\subsubsection{Temporal Alignment}
\label{methods:temporal_alignment}
Each sensing channel (modality) provides samples
$r^{(c)}_i$ at timestamps $\tau^{(c)}_i$, where $c \in \{\mathrm{P}, \mathrm{T}, \mathrm{US1}, \mathrm{US4}\}$. The pressure and temperature channels share
irregular timestamps at, on average, $\approx 7.5\,\mathrm{Hz}$, since the acquisition
software stores more samples during the filling phase, while the ultrasound channels are
exported as normalized RMS values at a regular $\approx 47.5\,\mathrm{Hz}$.
 We therefore align all channels on a uniform
time grid matched to the slower pressure/temperature recording. The grid is
\begin{equation}
  t_k = t_{\mathrm{start}} + k\,\Delta t, \qquad
  \Delta t = \frac{t_{\mathrm{end}} - t_{\mathrm{start}}}{T-1},
\end{equation}
for $k = 0,\dots,T-1$, where $[t_{\mathrm{start}}, t_{\mathrm{end}}]$ is the recording window covered
by all sensors,
\begin{gather}
  t_{\mathrm{start}} = \max_{c}\, \min_{i}\, \tau^{(c)}_i, \qquad
  t_{\mathrm{end}} = \min_{c}\, \max_{i}\, \tau^{(c)}_i, \\
  T = \bigl|\{\, i : t_{\mathrm{start}} \le \tau^{(\mathrm{P,T})}_i \le t_{\mathrm{end}} \,\}\bigr|,
\end{gather}
where $\tau^{(\mathrm{P,T})}_i$ are the timestamps
shared by the pressure and temperature channels.
The pressure/temperature recording spans $0.09$--$57.2\,\mathrm{s}$ and the
ultrasound recording $0.11$--$45.4\,\mathrm{s}$, so the pressure/temperature
signals are truncated from 403 to $T = 340$ samples, dropping one
sample before $t_{\mathrm{start}}$ and the $62$ samples after
$t_{\mathrm{end}}$, which lie in the residual cooling phase. This gives
${t_{\mathrm{end}} - t_{\mathrm{start}} \approx 45.3\,\mathrm{s}}$, i.e.\
$\Delta t \approx 0.134\,\mathrm{s}$.
Both settings use this window and thus share the sequence length $T$.

At each grid point $t_k$, a channel's value is linearly interpolated between
its two neighboring samples $\tau^{(c)}_i \le t_k \le \tau^{(c)}_{i+1}$,
\begin{equation}
  \bar{x}^{(c)}_k = (1 - \nu_k)\, r^{(c)}_i + \nu_k\, r^{(c)}_{i+1},
  \quad
  \nu_k = \frac{t_k - \tau^{(c)}_i}{\tau^{(c)}_{i+1} - \tau^{(c)}_i},
\end{equation}
and each channel is then min-max normalized to $[0,1]$,
\begin{equation}
  x^{(c)}_k = \frac{\bar{x}^{(c)}_k - \min_j \bar{x}^{(c)}_j}{\max_j \bar{x}^{(c)}_j - \min_j \bar{x}^{(c)}_j}.
\end{equation}

\subsubsection{Phase Extraction}
\label{methods:phase extraction}
We extract three phase markers from the normalized pressure $p_k = x^{(\mathrm{P})}_k$, namely the onset of the filling phase, the pressure maximum, and the end of the holding phase.
These markers are shared by all channels, since all channels lie on the same grid (Fig.~\ref{fig:modalities}).
With a fixed threshold $\theta = 0.1$, they are
\begin{align}
  k_{\mathrm{peak}} &= \arg\max_{k}\, p_k, \label{eq:k_peak}\\
  k_{\mathrm{fill}} &= \min\{\, k < k_{\mathrm{peak}} \;:\; p_k > \theta \,\}, \label{eq:k_fill}\\
  k_{\mathrm{hold}} &= \min\{\, k \ge k_{\mathrm{peak}} \;:\; p_k < \theta \,\}. \label{eq:k_hold}
\end{align}
The markers define three phases: the filling phase $\mathcal{F} = [k_{\mathrm{fill}}, k_{\mathrm{peak}})$; the holding phase $\mathcal{H} = [k_{\mathrm{peak}}, k_{\mathrm{hold}})$; and the cooling phase $\mathcal{C} = [k_{\mathrm{hold}}, T)$.

For the multi-modal failure modes, we additionally extract modality-specific indices, namely the temperature peak and the ultrasonic minima,
\begin{align}
  k^{(\mathrm{T})}_{\max} &= \arg\max_{k}\, x^{(\mathrm{T})}_k, \label{eq:k_T}\\
  k^{(c)}_{\min} &= \arg\min_{k}\, x^{(c)}_k, \qquad c \in \{\mathrm{US1}, \mathrm{US4}\}. \label{eq:k_us}
\end{align}
For the reference cycle, the markers take the following values, with times measured from $t_{\mathrm{start}}$:
\par\vspace{4pt}\noindent{\footnotesize\setlength{\tabcolsep}{0pt}%
\begin{tabular*}{\columnwidth}{@{\extracolsep{\fill}}lcccccc@{}}
    \toprule
    & $k_{\mathrm{fill}}$ & $k_{\mathrm{peak}}$ & $k_{\mathrm{hold}}$ & $k^{(\mathrm{T})}_{\max}$ & $k^{(\mathrm{US1})}_{\min}$ & $k^{(\mathrm{US4})}_{\min}$ \\
    \midrule
    Index & 25 & 28 & 114 & 33 & 28 & 144 \\
    Time [s] & 3.3 & 3.7 & 15.2 & 4.4 & 3.7 & 19.3 \\
    \bottomrule
\end{tabular*}\par}\vspace{4pt}

\subsubsection{Single-Modal Data}
\label{methods:single}
Part quality, and thus the decision to accept a part, depends strongly on the process parameters, among them the cavity pressure and its evolution throughout the cycle \cite{GimCavityPressAnalysis,https://doi.org/10.1155/2020/7023616}.

\begin{figure}[!t]
    \centering
    \includegraphics[width=\columnwidth]
        {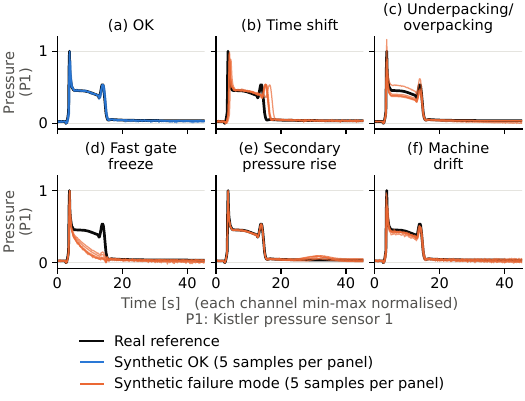}
    \caption{Synthetic single-modal data, one panel per class: (a) OK; (b) time shift, which moves the events later or earlier; (c) underpacking/overpacking, which scales the pressure over the filling and holding phases; (d) fast gate freeze, which decays the pressure over the holding phase; (e) secondary pressure rise, which adds a bump in the cooling phase; and (f) machine drift, which lowers the holding pressure and adds noise. Each panel shows five cycles of its class, OK in blue and failure modes in orange, over the real reference in black.}
    \label{fig:synthetic_pressure_only}
\end{figure}

\textit{OK disturbances} (Fig.~\ref{fig:synthetic_pressure_only}(a)).
For each cycle, we apply one to three distinct types from Table~\ref{tab:ok_disturbances}, chosen at random.

\begin{table}[!t]
    \centering
    \caption{OK Disturbances}
    \label{tab:ok_disturbances}
    \renewcommand{\arraystretch}{1.3}
    \footnotesize
    \begin{tabularx}{\columnwidth}{@{}>{\raggedright\arraybackslash}Xll@{}}
        \toprule
        Type & Operation & Range \\
        \midrule
        Gaussian noise & $x^{(c)}_k + \mathcal{N}(0, \sigma_{\mathrm{n}}^2)$ & $\sigma_{\mathrm{n}} \in [0.002, 0.01]$ \\
        Lead or lag    & $x^{(c)}_{k-\ell}$                            & $|\ell| \le 2$ \\
        Gain           & $f\, x^{(c)}_k$                               & $f \in [0.98, 1.02]$ \\
        Linear trend   & $x^{(c)}_k \pm m T k/(T-1)$                   & $m \in [2\cdot10^{-5}, 10^{-4}]$ \\
        Moving average & $\frac{1}{w}\sum_{j=0}^{w-1} x^{(c)}_{k-1+j}$ & $w \in \{2, 3\}$ \\
        \bottomrule
    \end{tabularx}
    \par\vspace{3pt}{\scriptsize\renewcommand{\arraystretch}{1}\raggedright Benign process variations applied to every channel $x^{(c)}$ of every synthetic cycle. One to three types are drawn per cycle, with parameters from the given ranges. Lead or lag and moving average pad with the edge value.\par}
\end{table}

\textit{Time shift} (Fig.~\ref{fig:synthetic_pressure_only}(b)).
We delay or advance the pressure profile by warping the normalized time axis $u_k = k/(T-1)$, while the overall signal shape stays intact.
We move each sample from $u_k$ to
\begin{equation}
    v_k = u_k + \alpha \sin(\pi u_k),
    \label{eq:warp}
\end{equation}
with shift fraction $|\alpha| \in [0.03, 0.07]$ of random sign, and linearly interpolate the points $(v_k, p_k)$ back onto the grid $\{u_k\}$.
The warp keeps the cycle start and end fixed; $\alpha > 0$ delays events and $\alpha < 0$ advances them.
The timing of the process state points, especially the end of the packing and cooling stages, is known to strongly influence part weight \cite{GimCavityPressAnalysis,polym15153207}.

\textit{Underpacking/overpacking} (Fig.~\ref{fig:synthetic_pressure_only}(c)).
We scale the pressure over the filling and holding phases,
\begin{equation}
    \tilde{p}_k = \phi\, p_k, \qquad k \in \mathcal{F} \cup \mathcal{H},
\end{equation}
with $\phi$ drawn from $[0.80, 0.92]$ (underpacking) with probability $0.6$ and from $[1.08, 1.20]$ (overpacking) otherwise.
Such variations in the packing pressure profile are known to drive defects like short shots or flash \cite{polym15153207,MOURYA2023656}.

\textit{Fast gate freeze} (Fig.~\ref{fig:synthetic_pressure_only}(d)).
We decay the pressure exponentially over the holding phase,
\begin{equation}
    \tilde{p}_k = p_k \exp\!\left(-\delta\, \frac{k - k_{\mathrm{peak}}}{|\mathcal{H}|}\right), \qquad k \in \mathcal{H},
\end{equation}
with decay rate $\delta \in [1, 2.5]$ and $|\mathcal{H}| = k_{\mathrm{hold}} - k_{\mathrm{peak}}$.
A fast pressure decay after the peak mimics premature gate freeze, since holding pressure can no longer push material into the cavity once the gate has frozen \cite{Zhao2022}.

\textit{Secondary pressure rise} (Fig.~\ref{fig:synthetic_pressure_only}(e)).
We add a Gaussian-shaped pressure bump after an onset sample $k_0 = \lfloor \xi T \rfloor$ with $\xi \in [0.40, 0.65]$,
\begin{equation}
    \tilde{p}_k = p_k + A \exp\!\left(-\frac{(k - k_0 - 0.3T_{\mathrm{r}})^2}{2\,(0.15T_{\mathrm{r}})^2}\right), \qquad k \ge k_0,
\end{equation}
where $T_{\mathrm{r}} = T - k_0$ is the number of remaining samples and $A \in [0.03, 0.10]$ is the amplitude.

\textit{Machine drift} (Fig.~\ref{fig:synthetic_pressure_only}(f)).
We model drift with a drift fraction $d \in [0.10, 0.35]$ that reduces the holding pressure,
\begin{equation}
    \tilde{p}_k = (1 - 0.4d)\, p_k, \qquad k \in \mathcal{H},
    \label{eq:drift}
\end{equation}
and adds broadband noise $\varepsilon_k \sim \mathcal{N}\bigl(0, (0.03d)^2\bigr)$ to every sample.
Ambient conditions and machine wear are known to cause such a drift of the process data during production \cite{yaninj}.

\subsubsection{Multi-Modal Data}
\label{methods:multi}
\begin{figure*}[!t]
    \centering
    \includegraphics[width=\textwidth]
        {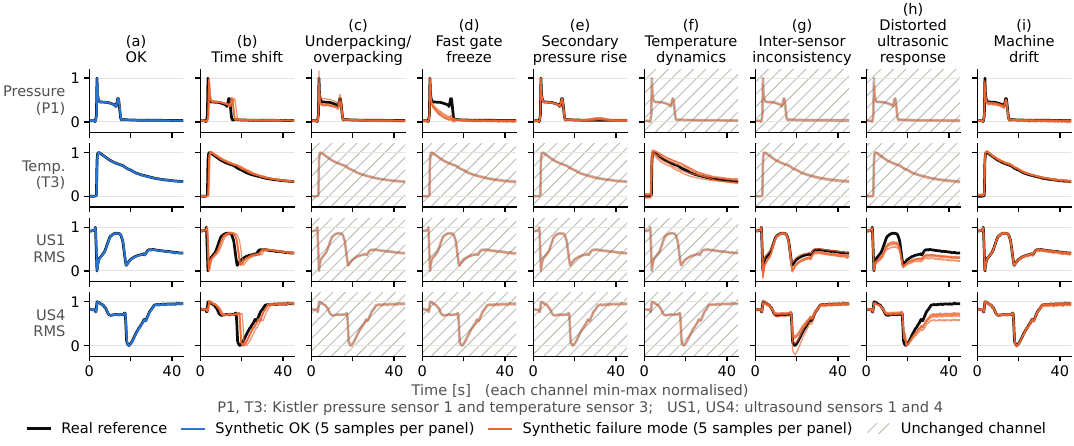}
    \caption{Synthetic multi-modal data, one column per class: (a) OK, (b) time shift, (c) underpacking/overpacking, (d) fast gate freeze, (e) secondary pressure rise, (f) temperature dynamics, (g) inter-sensor inconsistency, (h) distorted ultrasonic response, and (i) machine drift. Rows show the pressure P1, temperature T3, and ultrasound US1 and US4 channels. Time shift and machine drift act on all channels, modes (c)--(e) only on the pressure, (f) only on the temperature, and (g) and (h) only on the ultrasound. Each panel shows five synthetic cycles, OK in blue and failure modes in orange, over the real reference in black. Hatched, faded panels mark channels that the failure mode leaves unchanged; these differ from the reference only by the OK disturbances applied to every cycle.}
    \label{fig:synthetic_multi-modal}
\end{figure*}

For multi-modal failure modes, we keep the underpacking/overpacking, fast gate freeze, and secondary pressure rise of the single-modal setting (Section~\ref{methods:single}) unchanged (Fig.~\ref{fig:synthetic_multi-modal}(c)--(e)), adapt the time shift and machine drift modes, and introduce three modes that change only the additional modalities.
The OK disturbances act on every channel $x^{(c)}$ (Table~\ref{tab:ok_disturbances}).

\textit{Time shift} (Fig.~\ref{fig:synthetic_multi-modal}(b)).
We apply the warp~\eqref{eq:warp} with the same $\alpha$ to every channel $x^{(c)}$.
Since this failure mode changes the phase timing of the process, all modalities reflect it.

\textit{Machine drift} (Fig.~\ref{fig:synthetic_multi-modal}(i)).
In addition to the pressure drift~\eqref{eq:drift}, we raise the temperature baseline, $\tilde{x}^{(\mathrm{T})}_k = x^{(\mathrm{T})}_k + 0.15d$, and the broadband noise $\varepsilon_k$ now also perturbs the temperature and ultrasonic channels.
The two perturbations are causally independent, and the temperature rise is a modeling choice.
Elevated melt temperature is known to promote material degradation and excessive post-cooling shrinkage \cite{Zhao2022}.

\textit{Inter-sensor inconsistency} (Fig.~\ref{fig:synthetic_multi-modal}(g)).
We rescale both ultrasonic RMS channels around their pre-fill level $x^{(c)}_0$,
\begin{equation}
    \tilde{x}^{(c)}_k = x^{(c)}_0 + \gamma_c\bigl(x^{(c)}_k - x^{(c)}_0\bigr), \qquad c \in \{\mathrm{US1}, \mathrm{US4}\},
\end{equation}
where one factor is drawn from $[0.75, 0.95]$ and the other from $[1.05, 1.25]$, assigned to the sensors at random, so that one sensor is attenuated while the other is amplified over the entire cycle.
We use this mode to model a difference in contact between the two sensor locations, since ultrasonic amplitude reflects the contact between the cavity wall and the polymer \cite{kariminejad2021USinIM}. Such a difference can indicate uneven shrinkage, which is known to cause warpage \cite{Zhao2022}. 

\textit{Distorted ultrasonic response} (gap-formation hypothesis; Fig.~\ref{fig:synthetic_multi-modal}(h)).
We rescale each ultrasonic channel in two periods, split at its minimum $k^{(c)}_{\min}$ from~\eqref{eq:k_us}, with a dip scale $\rho_{\mathrm{d}}$ and a recovery scale $\rho_{\mathrm{r}}$, both drawn from $[0.60, 0.90]$:
\begin{equation}
    \tilde{x}^{(c)}_k =
    \begin{cases}
        x^{(c)}_0 + \rho_{\mathrm{d}}\bigl(x^{(c)}_k - x^{(c)}_0\bigr), & k_{\mathrm{peak}} \le k < k^{(c)}_{\min},\\[2pt]
        x^{(c)}_{k^{(c)}_{\min}} + \rho_{\mathrm{r}}\bigl(x^{(c)}_k - x^{(c)}_{k^{(c)}_{\min}}\bigr), & k \ge k^{(c)}_{\min}.
    \end{cases}
\end{equation}
Between the pressure peak and the minimum, we pull the signal towards its pre-fill level, which reduces its drop before the minimum. From the minimum onward, we pull it towards the minimum value, which suppresses the recovery. The minimum value itself stays unchanged.
For $x^{(\mathrm{US1})}$, the minimum coincides with the pressure peak, so only the recovery changes. For $x^{(\mathrm{US4})}$, it lies in the cooling phase, so the first period spans the holding phase and the beginning of the cooling phase.
Unlike inter-sensor inconsistency, this mode changes both sensors in the same direction, from the pressure peak onward. We use it to model a gap forming at both sensor locations, since changes in ultrasonic amplitude and in the reflection and transmission at the wall reflect gap formation and part detachment \cite{kariminejad2021USinIM}.

\textit{Temperature dynamics} (Fig.~\ref{fig:synthetic_multi-modal}(f)).
First, we shift the whole temperature channel, $\check{x}^{(\mathrm{T})}_k = x^{(\mathrm{T})}_k + b$, with $|b| \in [0.02, 0.06]$ of random sign, to emulate a changed bulk melt temperature \cite{polym15214220}, where $b > 0$ raises the risk of material degradation and shrinkage \cite{Zhao2022}.
Second, we warp the cooling segment from the temperature peak $k^{(\mathrm{T})}_{\max}$ from~\eqref{eq:k_T} to the end of the cycle.
With $T_{\mathrm{c}} = T - k^{(\mathrm{T})}_{\max}$ and $u_j = j/(T_{\mathrm{c}}-1)$, $j = 0, \dots, T_{\mathrm{c}}-1$, the segment becomes
\begin{equation}
    \tilde{x}^{(\mathrm{T})}_{k^{(\mathrm{T})}_{\max}+j} = \check{x}^{(\mathrm{T})}\bigl(u_j^{1/\kappa}\bigr),
\end{equation}
where $\check{x}^{(\mathrm{T})}(u)$ is the linear interpolation of the shifted segment at normalized time $u \in [0, 1]$.
The peak and the last value stay in place, and only the temperature between them changes: $\kappa$ is drawn with equal probability from $[1.1, 1.4]$, where the temperature falls earlier (faster cooling), or from $[0.7, 0.9]$, where it falls later (slower cooling).
Cooling rate is known to influence crystallization kinetics and crystal size in injection-moulded polymers \cite{10.1115/IMECE2021-73665}.

\subsection{MouldIt Model}
\label{subsec:mouldit}
\begin{figure*}[!t]
    \centering
    \includegraphics[width=0.96\textwidth]
        {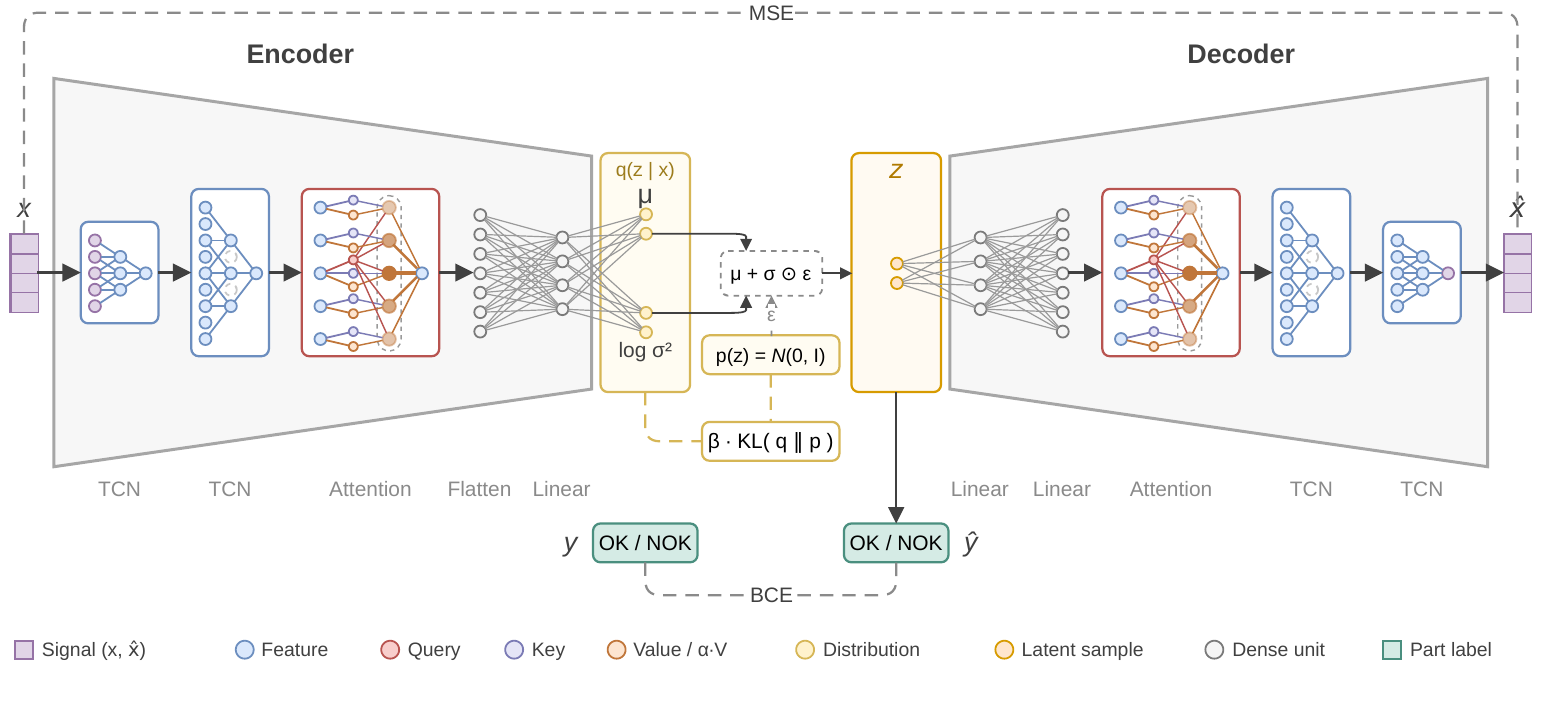}
    \caption{MouldIt architecture. The input $\mathbf{x}$ has $C$ channels of length $T = 340$, with $C = 1$ in the single-modal and $C = 4$ in the multi-modal setting. The encoder consists of two TCN blocks, a temporal attention block, and a flatten and a linear layer, and predicts the mean $\boldsymbol{\mu}$ and log variance $\log\boldsymbol{\sigma}^2$ of the 32-dimensional latent distribution $q(\mathbf{z} \mid \mathbf{x})$. A latent sample $\mathbf{z} = \boldsymbol{\mu} + \boldsymbol{\sigma} \odot \boldsymbol{\epsilon}$, with $\boldsymbol{\epsilon} \sim \mathcal{N}(\mathbf{0}, \mathbf{I})$, feeds two branches: $Q$ linear heads predict the part label $\hat{y}$ (OK or a NOK variant), and a decoder of two linear layers, a temporal attention block, and two TCN blocks reconstructs the signal $\hat{\mathbf{x}}$. Dashed lines mark the three loss terms: the MSE between $\mathbf{x}$ and $\hat{\mathbf{x}}$, the $\beta$-weighted KLD between $q$ and the prior $p(\mathbf{z})$, and the BCE between $y$ and $\hat{y}$. At inference, $\boldsymbol{\mu}$ replaces $\mathbf{z}$.}
    \label{fig:mouldit_model}
\end{figure*}

We formulate MouldIt as a variational autoencoder, with $Q$ additional prediction heads, one per known class, projected directly from the encoder embedding (Fig.~\ref{fig:mouldit_model}). The encoder comprises two temporal convolution blocks to capture local features, followed by a temporal attention block to model long-range dependencies between the captured features. We flatten the result and pass it through a linear layer followed by two linear heads to predict the $\boldsymbol{\mu}$ and $\log\boldsymbol{\sigma}^2$ of the latent distribution. After resampling, we pass the latent sample in parallel to two different network blocks. First, a block of $Q$ linear heads projects the latent sample to the probabilities of OK and each NOK variant. Second, a decoder mirrors the encoder architecture in reverse, with a small variation in the channel widths of the temporal convolution blocks.

MouldIt optimizes a loss composed of three components~\eqref{eq:loss}: a reconstruction MSE
between the input signal and the reconstructed signal, a KLD regularizer
between the approximate posterior and a standard normal prior, and a BCE
between the ground-truth part labels and their predicted probabilities.
\begin{equation}
  \mathcal{L} = \lambda_{\mathrm{rec}} \left( \mathcal{L}_{\mathrm{MSE}}
              + \beta(e)\, \mathcal{L}_{\mathrm{KLD}} \right)
              + \lambda_{\mathrm{cls}}\, \mathcal{L}_{\mathrm{BCE}},
  \label{eq:loss}
\end{equation}
where $e$ is the training epoch. The first two loss terms together are weighted against the BCE ($\lambda_{\mathrm{rec}} = 1$ and $\lambda_{\mathrm{cls}} = 0.1$).

For a batch of $B$ signals $\mathbf{x}_i \in \mathbb{R}^{C \times T}$ with
$C \in \{1, 4\}$ channels and sequence length $T$ (Section~\ref{methods:temporal_alignment}), and their reconstructions
$\hat{\mathbf{x}}_i$, the reconstruction MSE is
\begin{equation}
  \mathcal{L}_{\mathrm{MSE}} = \frac{1}{BCT} \sum_{i=1}^{B} \sum_{c=1}^{C}
  \sum_{t=1}^{T} \left( x_{i,c,t} - \hat{x}_{i,c,t} \right)^2 .
\end{equation}
The approximate posterior is a diagonal Gaussian
$q(\mathbf{z} \mid \mathbf{x}_i) = \mathcal{N}\!\left(\boldsymbol{\mu}_i, \operatorname{diag}(\boldsymbol{\sigma}_i^2)\right)$
over $D = 32$ dimensions, where $\boldsymbol{\mu}_i$ and $\log \boldsymbol{\sigma}_i^2$ are predicted by the encoder, and its KLD against the standard normal prior
$\mathcal{N}(\mathbf{0}, \mathbf{I})$ is
\begin{equation}
  \mathcal{L}_{\mathrm{KLD}} = -\frac{1}{2B} \sum_{i=1}^{B} \sum_{j=1}^{D}
  \left( 1 + \log \sigma_{i,j}^2 - \mu_{i,j}^2 - \sigma_{i,j}^2 \right).
\end{equation}
The KLD is weighted by $\beta(e)$:
\begin{equation}
  \beta(e) = \beta \min\!\left(1, \frac{e}{100}\right),
  \quad e = 0, 1, 2, \dots
\end{equation}
We set $\beta = 2\sigma_{\mathrm{OK}}^2 / (CT)$, where $CT$ is the number of values in one cycle and $\sigma_{\mathrm{OK}}$ is the root mean square deviation of synthetic OK cycles from the reference cycle, estimated once on 300 OK cycles. With this choice, the first two loss terms match, up to a constant factor, those of a VAE whose Gaussian decoder assumes the noise variance $\sigma_{\mathrm{OK}}^2$. With $\sigma_{\mathrm{OK}} = 0.0304$ and $0.0253$, this gives $\beta = 5.4 \times 10^{-6}$ in the single-modal and $\beta = 9.4 \times 10^{-7}$ in the multi-modal setting, which we fix for all seeds and dataset sizes.

Each of the $Q$ heads predicts one class in a one-vs-rest manner. The BCE between the one-hot labels $y_{iq} \in \{0,1\}$ and the predicted
class probabilities $\hat{y}_{iq} \in (0,1)$ is
\begin{equation}
  \mathcal{L}_{\mathrm{BCE}} = -\frac{1}{BQ} \sum_{i=1}^{B} \sum_{q=1}^{Q}
  \bigl[ y_{iq} \log \hat{y}_{iq} + (1 - y_{iq}) \log(1 - \hat{y}_{iq}) \bigr],
\end{equation}
and the predicted class is the one with the highest probability. The part quality prediction problem thus maps $\mathbb{R}^{C \times T} \to \mathbb{R}^{Q}$, while the reconstruction returns the same input space.

\subsection{Novelty Score}
\label{subsec:novelty}
To detect unknown faults, we compute a novelty score from the outputs of the trained MouldIt model, without changing the network. For every cycle $\mathbf{x}$, it combines a latent term, from its position in the latent space, and a reconstruction term, from its reconstruction error.
For the latent term, we fit one Gaussian per known class $q$ to the training cycles in the latent space, each represented by its predicted mean $\boldsymbol{\mu}$, using Ledoit--Wolf shrinkage \cite{ledoit2004wellconditioned}.
With $M_q(\mathbf{x})$ the squared Mahalanobis distance of $\boldsymbol{\mu}$ to class $q$, and $\eta_q$ its 95th percentile on the validation cycles of that class, the latent term is the distance to the nearest class in units of its 95\% boundary,
\begin{equation}
  L(\mathbf{x}) = \min_{q}\, \frac{M_q(\mathbf{x})}{\eta_q}.
\end{equation}
For the reconstruction term, we use the summed squared error $R(\mathbf{x}) = \lVert \mathbf{x} - \hat{\mathbf{x}} \rVert^2$, with $\eta_R$ its 95th percentile on all known validation cycles.
The novelty score is
\begin{equation}
  N(\mathbf{x}) = \max\!\left( L(\mathbf{x}),\ \frac{R(\mathbf{x})}{\eta_R} \right).
  \label{eq:novelty}
\end{equation}
At inference, we use $\boldsymbol{\mu}$ instead of a sampled $\mathbf{z}$, so the score is deterministic.
PCA uses the same score with 32 components, the size of the MouldIt latent space: $L$ is computed on the PCA scores (Hotelling's $T^2$ per class) and $R$ is the squared prediction error (SPE). PCA predicts the class with the smallest $M_q / \eta_q$.
The remaining baselines have a single score: the mean distance to the five nearest standardized cycles for kNN, and one minus the highest class probability for logistic regression and random forests (200 trees).
Every method flags a cycle as unknown when its score exceeds $\Theta$, the 95th percentile of the score on the known validation cycles.

\begin{figure*}[t]
    \centering
    \includegraphics[width=0.93\textwidth]
        {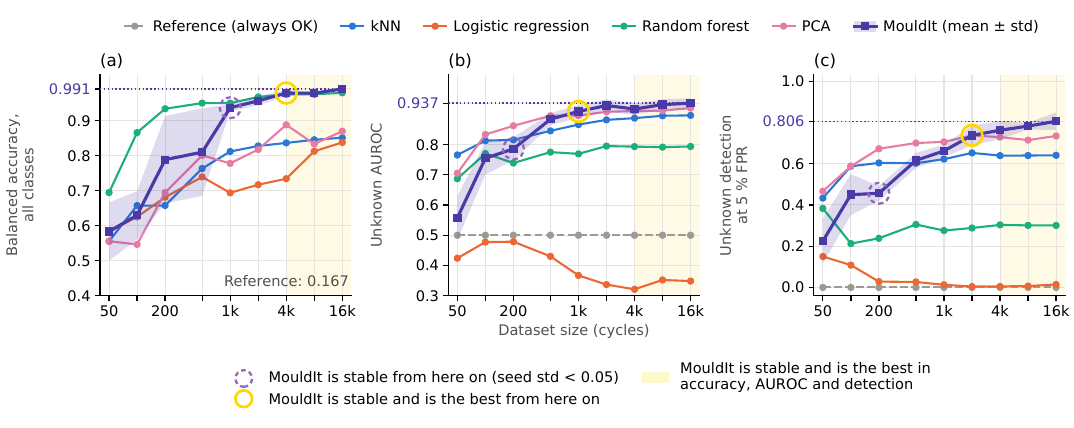}
    \caption{Single-modal results over dataset size: (a) balanced accuracy over all six classes, and (b) unknown AUROC and (c) unknown detection at a 5\% false positive rate, both for failure modes held out from training and averaged over the five modes. Lines show the mean over three seeds and the band the standard deviation of MouldIt; the dotted line and value mark MouldIt at 16000 cycles. The dashed grey line is the reference that accepts every part, with a chance-level accuracy of 1/6. Random forests with the PCA unknown flag are not shown, since they have no classifier or score of their own. Dashed circles mark the size from which MouldIt stays stable across seeds (standard deviation below 0.05), solid circles the size from which it is also the best method, and the shaded region the sizes where it is stable and the best in all three metrics, here from 4000 cycles.}
    \label{fig:scaling_pressure_only}
\end{figure*}

\section{Experiments}
\label{sec:experiments}
We compare MouldIt with classical methods on two tasks: classifying cycles into OK or one of the known failure modes, and detecting unknown failure modes that were not part of training. We are particularly interested in how performance depends on the amount of training data. For both the single-modal (pressure) and the multi-modal (pressure, temperature, ultrasound) setting, we therefore generate datasets of 50, 100, 200, 500, 1000, 2000, 4000, 8000, and 16000 cycles from the real reference cycle (Section~\ref{subsec:synthetic-data-generation}). All classes are equally represented, with six classes in the single-modal and nine in the multi-modal setting, so each size is rounded down to a multiple of the number of classes, for example to 15996 and 15993 cycles at 16000. We split each dataset into 60\% training, 20\% validation, and 20\% test data, balanced by class.

MouldIt (Section~\ref{subsec:mouldit}) is trained for 1000 epochs with cosine annealing from a learning rate of 0.001. As baselines, we use k-nearest neighbors (kNN), logistic regression, random forests, principal component analysis (PCA), and a trivial reference that always predicts OK. We run all of them on our data, including methods used in prior IM studies, such as random forests~\cite{polenta22MLinj} and kNN~\cite{mlcavitypressuredataefficiency}. We do not compare numerically with the published IM models of Section~\ref{sec:related}: they report results on other parts, sensors, and failure modes, and, to our knowledge, the closest one~\cite{yaninj} has no public implementation. Every run is repeated with three seeds, and we report the mean $\pm$ standard deviation.

\textbf{Classification.} Here, all classes are seen during training, and we report the balanced accuracy over all classes.

\textbf{Unknown detection.} \textit{Setup:} We hold out one failure mode at a time from training and validation. All of its cycles, the unknown faults, are added to the test set.

\textit{Novelty score:} Each method assigns a novelty score to every cycle (Section~\ref{subsec:novelty}). The threshold $\Theta$ corresponds to a 5\% false positive rate (FPR) on the known validation cycles.

\textit{Metrics:} We report the unknown AUROC of the novelty score and the unknown detection at 5\% FPR, i.e., the share of unknown faults that are flagged. Both are averaged over the held-out failure modes.

\textbf{Sorting losses.} A part is shipped only if it is not flagged and is predicted OK. This decision can go wrong in four ways: good parts rejected, known faults shipped, unknown faults shipped, and unknown faults rejected as a known fault. We call the mean of these four rates the sorting loss, report it over dataset size, and rank the methods at 16000 cycles. Besides the baselines above, we evaluate random forests with the PCA unknown flag, which ships a part only if random forests predict OK and PCA does not flag it.

\subsection{Single-Modal Experiments}
\label{sec:exp_single}
\begin{figure*}[t]
    \centering
    \includegraphics[width=0.91\textwidth]
        {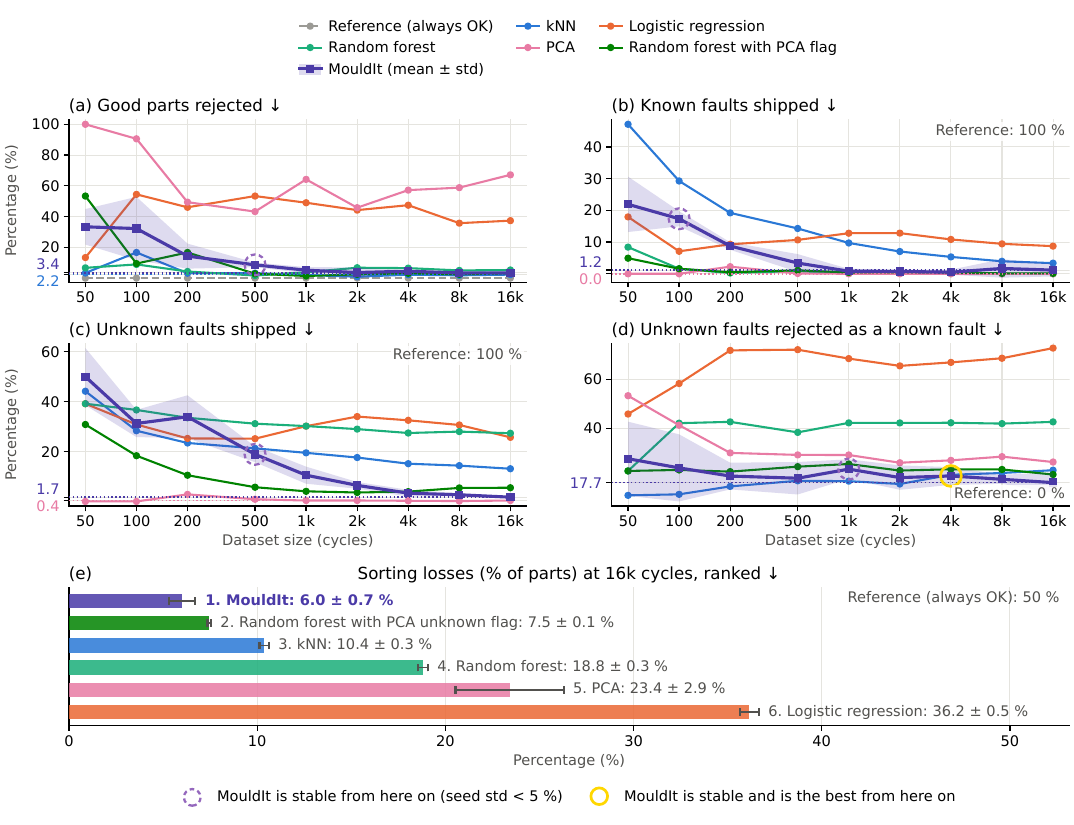}
    \caption{Single-modal sorting errors over dataset size: (a) good parts rejected, (b) known faults shipped, (c) unknown faults shipped, and (d) unknown faults rejected as a known fault, where a part is shipped only if it is predicted OK and not flagged as unknown. Lines show the mean over three seeds and the band the standard deviation of MouldIt. The reference accepts every part, so it rejects no good part and ships every fault. Dashed circles mark the size from which MouldIt stays stable across seeds (standard deviation below 5\%), and solid circles the size from which it is also the best method. Panel (e) ranks the methods by their sorting loss, the mean of the four rates, at 16000 cycles.}
    \label{fig:sorting_pressure_only}
\end{figure*}
\subsubsection{Classification}
With little training data, random forests are clearly the most accurate classifier, reaching a balanced accuracy of 86.6 $\pm$ 2.1\% at a size of 100 cycles (Fig.~\ref{fig:scaling_pressure_only}(a)). MouldIt's balanced accuracy becomes consistent across seeds from a size of 1000, and from 4000 onward it is higher than that of random forests. At 16000, MouldIt reaches 99.1 $\pm$ 0.5\% compared with 97.9 $\pm$ 0.3\% for random forests. None of the other baselines exceeds 88\%.

\subsubsection{Unknown Detection}
When training data is scarce, PCA is the best detector. MouldIt overtakes it in unknown AUROC from a size of 1000 (Fig.~\ref{fig:scaling_pressure_only}(b)). In unknown detection, it ties with PCA at 2000 and leads from 4000 onward (Fig.~\ref{fig:scaling_pressure_only}(c)). At 16000, MouldIt reaches an unknown AUROC of 0.937 $\pm$ 0.016 and detects 80.6 $\pm$ 4.2\% of the unknown faults at 5\% FPR. PCA follows with 0.921 $\pm$ 0.004 and 73.5 $\pm$ 0.4\%. Machine drift is the hardest failure mode for MouldIt, with an unknown AUROC of 0.823, since it deviates from the reference only slightly more than normal OK cycles do. For PCA, the hardest failure mode is underpacking/overpacking, with 0.766.

\subsubsection{Sorting Losses}
With little training data, random forests with the PCA unknown flag sort parts best, with a sorting loss of 12.5 $\pm$ 3.2\% at a size of 200 (Fig.~\ref{fig:sorting_pressure_only}). MouldIt catches up at 2000 and has the lowest sorting loss from 4000 onward. At 16000, its sorting loss is 6.0 $\pm$ 0.7\% (Fig.~\ref{fig:sorting_pressure_only}(e)), followed by random forests with the PCA unknown flag with 7.5 $\pm$ 0.1\% and kNN with 10.4 $\pm$ 0.3\%. The remaining methods have sorting losses between 18.8\% and 36.2\%, and the reference has 50.0\%. Most of MouldIt's sorting loss comes from unknown faults rejected as a known fault (17.7 $\pm$ 4.2\%; Fig.~\ref{fig:sorting_pressure_only}(d)); its other three rates stay below 4\% (Fig.~\ref{fig:sorting_pressure_only}(a)--(c)). PCA rejects 67.1 $\pm$ 11.1\% of the good parts (Fig.~\ref{fig:sorting_pressure_only}(a)), and random forests ship 27.3 $\pm$ 2.0\% of the unknown faults (Fig.~\ref{fig:sorting_pressure_only}(c)).

\subsection{Multi-Modal Experiments}
\label{sec:exp_multi}
\begin{figure*}[t]
    \centering
    \includegraphics[width=0.93\textwidth]
        {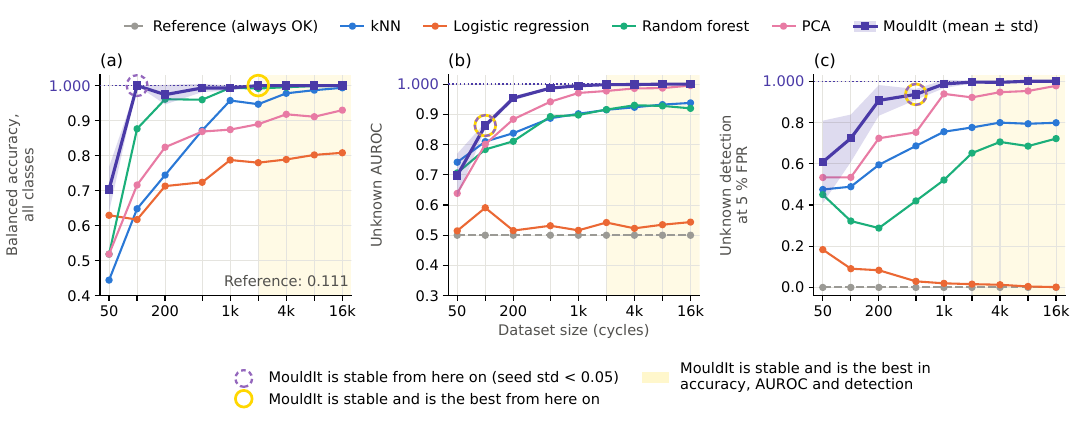}
    \caption{Multi-modal results over dataset size: (a) balanced accuracy over all nine classes, and (b) unknown AUROC and (c) unknown detection at a 5\% false positive rate, both for failure modes held out from training and averaged over the eight modes. Lines show the mean over three seeds and the band the standard deviation of MouldIt; the dotted line and value mark MouldIt at 16000 cycles. The dashed grey line is the reference that accepts every part, with a chance-level accuracy of 1/9. Random forests with the PCA unknown flag are not shown, since they have no classifier or score of their own. Dashed circles mark the size from which MouldIt stays stable across seeds (standard deviation below 0.05), solid circles the size from which it is also the best method, and the shaded region the sizes where it is stable and the best in all three metrics, here from 2000 cycles.}
    \label{fig:scaling_multi-modal}
\end{figure*}
\begin{figure*}[t]
    \centering
    \includegraphics[width=0.91\textwidth]
        {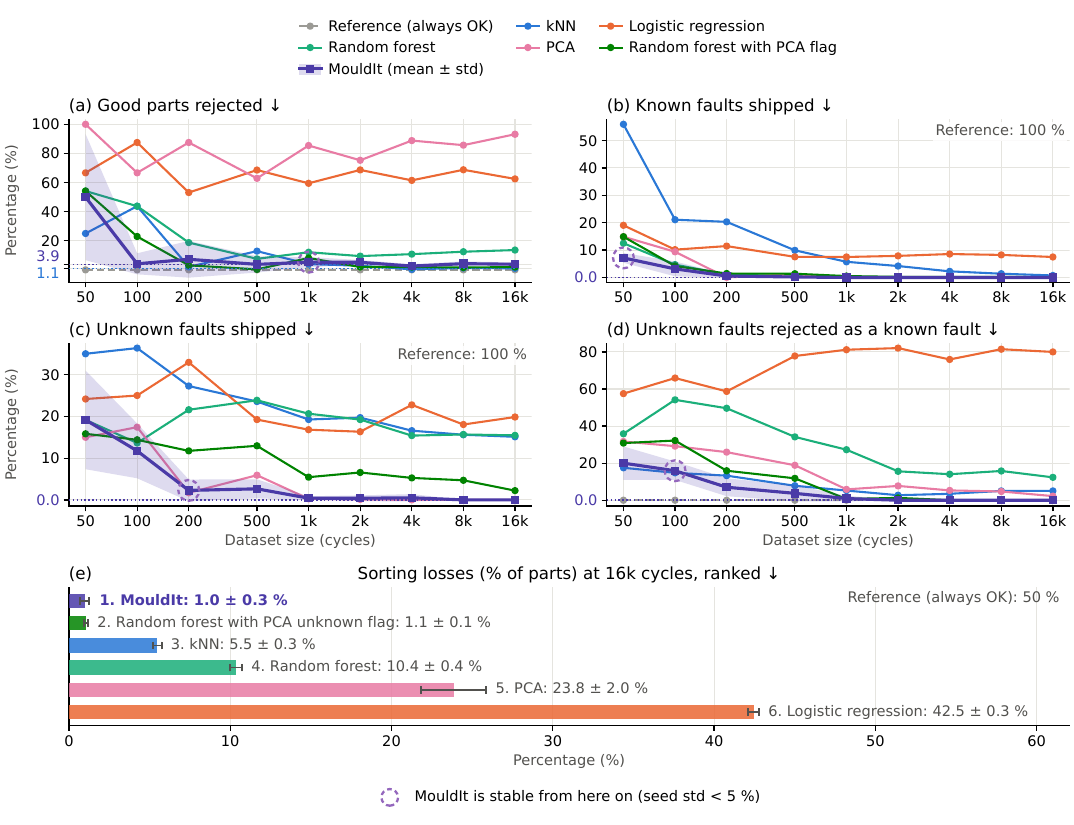}
    \caption{Multi-modal sorting errors over dataset size: (a) good parts rejected, (b) known faults shipped, (c) unknown faults shipped, and (d) unknown faults rejected as a known fault, where a part is shipped only if it is predicted OK and not flagged as unknown. Lines show the mean over three seeds and the band the standard deviation of MouldIt. The reference accepts every part, so it rejects no good part and ships every fault. Dashed circles mark the size from which MouldIt stays stable across seeds (standard deviation below 5\%). Panel (e) ranks the methods by their sorting loss, the mean of the four rates, at 16000 cycles.}
    \label{fig:sorting_multi-modal}
\end{figure*}
\subsubsection{Classification}
With multi-modal data, MouldIt reaches a balanced accuracy at least as high as every baseline at all dataset sizes (Fig.~\ref{fig:scaling_multi-modal}(a)). From a size of 2000 onward, it classifies every test cycle correctly in all seeds. At 16000, random forests and kNN come close with 99.9 $\pm$ 0.1\% and 99.4 $\pm$ 0.3\%, whereas PCA and logistic regression stay below 94\%.

\subsubsection{Unknown Detection}
MouldIt has the highest unknown detection at every dataset size (Fig.~\ref{fig:scaling_multi-modal}(c)). Already at a size of 200, it flags 90.7 $\pm$ 7.4\% of the unknown faults, compared with 72.3 $\pm$ 5.1\% for PCA. From 8000 onward, it flags all unknown faults in every seed. At 16000, PCA comes closest with an unknown AUROC of 0.995 $\pm$ 0.000 (Fig.~\ref{fig:scaling_multi-modal}(b)) and an unknown detection of 97.8 $\pm$ 0.2\%. None of the other baselines detects more than 80\%.

\subsubsection{Sorting Losses}
MouldIt has the lowest sorting loss at every dataset size (Fig.~\ref{fig:sorting_multi-modal}). At a size of 200, its sorting loss is 4.3 $\pm$ 1.9\%, roughly half that of random forests with the PCA unknown flag (8.1 $\pm$ 1.3\%). At 16000, MouldIt's sorting loss of 1.0 $\pm$ 0.3\% (Fig.~\ref{fig:sorting_multi-modal}(e)) is on par with that of random forests with the PCA unknown flag (1.1 $\pm$ 0.1\%). The other baselines have higher sorting losses: 5.5 $\pm$ 0.3\% for kNN, 10.4 $\pm$ 0.4\% for random forests, 23.8 $\pm$ 2.0\% for PCA, and 42.5 $\pm$ 0.3\% for logistic regression. MouldIt's only errors are good parts rejected (3.9 $\pm$ 1.2\%; Fig.~\ref{fig:sorting_multi-modal}(a)).

\section{Discussion}
\label{sec:discussion}
We discuss the data efficiency and multi-modal benefit of MouldIt, how it detects unknown faults, how well it reconstructs them, and the limitations of our study.

\subsection{Data Efficiency and Multi-Modal Benefit}
Comparing the two settings shows that additional modalities reduce the data MouldIt needs. With pressure alone (Figs.~\ref{fig:scaling_pressure_only} and~\ref{fig:sorting_pressure_only}), it has not fully saturated even at 16000 cycles: between 4000 and 16000 cycles, its unknown detection still rises by 4.4 percentage points and its sorting loss falls by 1.2. With all four channels (Fig.~\ref{fig:sorting_multi-modal}), its sorting loss no longer improves by more than the seed variation from 4000 cycles. The benefit is largest with little data: at 200 cycles, the multi-modal sorting loss is 4.5 times lower than the single-modal one.

The two settings are not the same problem with more sensors. The multi-modal setting has more classes, and three of its failure modes exist only in the added channels (Section~\ref{methods:multi}). The classical methods also profit from these channels (Table~\ref{tab:metrics}), so the benefit is not specific to MouldIt. What sets MouldIt apart is that, with all four channels, it needs less data for a low sorting loss: it falls below 5\% at 200 cycles, whereas random forests with the PCA unknown flag need 1000 (Fig.~\ref{fig:sorting_multi-modal}). In addition, one jointly trained model does the work of two separate ones. MouldIt does not lead on every sorting error, but it has the lowest overall sorting loss (Table~\ref{tab:metrics}).

\begin{table*}[!t]
    \centering
    \caption{Results at 16000 Cycles}
    \label{tab:metrics}
    \renewcommand{\arraystretch}{1.08}
    \footnotesize
    \setlength{\tabcolsep}{4pt}
    \begin{tabular*}{\textwidth}{@{\extracolsep{\fill}}lcccccccc@{}}
        \toprule
        & \multicolumn{3}{c}{Higher is better} & \multicolumn{5}{c}{Lower is better} \\
        \cmidrule(lr){2-4} \cmidrule(l){5-9}
        Method & \begin{tabular}[c]{@{}c@{}}Balanced\\accuracy\end{tabular} & \begin{tabular}[c]{@{}c@{}}Unknown\\AUROC\end{tabular} & \begin{tabular}[c]{@{}c@{}}Unknown\\detection\end{tabular} & \begin{tabular}[c]{@{}c@{}}Good parts\\rejected\end{tabular} & \begin{tabular}[c]{@{}c@{}}Known faults\\shipped\end{tabular} & \begin{tabular}[c]{@{}c@{}}Unknown faults\\shipped\end{tabular} & \begin{tabular}[c]{@{}c@{}}Unknown rejected\\as known\end{tabular} & \begin{tabular}[c]{@{}c@{}}Sorting\\loss\end{tabular} \\
        \midrule
\multicolumn{9}{@{}l}{\textit{Single-modal (pressure; 6 classes)}}\\
MouldIt & \textbf{99.1 $\pm$ 0.5} & \textbf{0.937 $\pm$ 0.016} & \textbf{80.6 $\pm$ 4.2} & 3.4 $\pm$ 1.3 & 1.2 $\pm$ 2.0 & 1.7 $\pm$ 0.2 & \textbf{17.7 $\pm$ 4.2} & \textbf{6.0 $\pm$ 0.7} \\
RF + PCA flag & -- & -- & -- & 3.2 $\pm$ 0.7 & 0.1 $\pm$ 0.1 & 5.5 $\pm$ 0.7 & 21.0 $\pm$ 0.4 & 7.5 $\pm$ 0.1 \\
kNN & 85.1 $\pm$ 0.1 & 0.897 $\pm$ 0.001 & 64.1 $\pm$ 0.6 & \textbf{2.2 $\pm$ 1.3} & 3.4 $\pm$ 0.2 & 13.1 $\pm$ 0.4 & 22.8 $\pm$ 0.2 & 10.4 $\pm$ 0.3 \\
Random forest & 97.9 $\pm$ 0.3 & 0.794 $\pm$ 0.011 & 30.1 $\pm$ 1.3 & 5.3 $\pm$ 0.3 & \textbf{0.0 $\pm$ 0.0} & 27.3 $\pm$ 2.0 & 42.6 $\pm$ 0.8 & 18.8 $\pm$ 0.3 \\
PCA & 87.0 $\pm$ 1.6 & 0.921 $\pm$ 0.004 & 73.5 $\pm$ 0.4 & 67.1 $\pm$ 11.1 & \textbf{0.0 $\pm$ 0.0} & \textbf{0.4 $\pm$ 0.3} & 26.2 $\pm$ 0.7 & 23.4 $\pm$ 2.9 \\
Logistic regression & 83.8 $\pm$ 0.9 & 0.348 $\pm$ 0.010 & 1.5 $\pm$ 0.4 & 37.3 $\pm$ 2.2 & 8.7 $\pm$ 0.2 & 25.7 $\pm$ 4.5 & 72.9 $\pm$ 4.1 & 36.2 $\pm$ 0.5 \\
Reference (always OK) & 16.7 $\pm$ 0.0 & 0.500 $\pm$ 0.000 & 0.0 $\pm$ 0.0 & 0.0 $\pm$ 0.0 & 100.0 $\pm$ 0.0 & 100.0 $\pm$ 0.0 & 0.0 $\pm$ 0.0 & 50.0 $\pm$ 0.0 \\
\midrule
\multicolumn{9}{@{}l}{\textit{Multi-modal (pressure, temperature, two ultrasound channels; 9 classes)}}\\
MouldIt & \textbf{100.0 $\pm$ 0.0} & \textbf{1.000 $\pm$ 0.000} & \textbf{100.0 $\pm$ 0.0} & 3.9 $\pm$ 1.2 & \textbf{0.0 $\pm$ 0.0} & \textbf{0.0 $\pm$ 0.0} & \textbf{0.0 $\pm$ 0.0} & \textbf{1.0 $\pm$ 0.3} \\
RF + PCA flag & -- & -- & -- & 1.9 $\pm$ 0.4 & 0.1 $\pm$ 0.1 & 2.2 $\pm$ 0.2 & \textbf{0.0 $\pm$ 0.0} & 1.1 $\pm$ 0.1 \\
kNN & 99.4 $\pm$ 0.3 & 0.938 $\pm$ 0.002 & 79.9 $\pm$ 0.4 & \textbf{1.1 $\pm$ 0.7} & 0.7 $\pm$ 0.2 & 15.1 $\pm$ 0.2 & 5.0 $\pm$ 0.5 & 5.5 $\pm$ 0.3 \\
Random forest & 99.9 $\pm$ 0.1 & 0.920 $\pm$ 0.008 & 72.2 $\pm$ 1.1 & 13.6 $\pm$ 0.7 & \textbf{0.0 $\pm$ 0.0} & 15.5 $\pm$ 0.5 & 12.3 $\pm$ 0.6 & 10.4 $\pm$ 0.4 \\
PCA & 93.0 $\pm$ 4.8 & 0.995 $\pm$ 0.000 & 97.8 $\pm$ 0.2 & 93.2 $\pm$ 8.3 & \textbf{0.0 $\pm$ 0.0} & \textbf{0.0 $\pm$ 0.0} & 2.2 $\pm$ 0.2 & 23.8 $\pm$ 2.0 \\
Logistic regression & 80.8 $\pm$ 0.8 & 0.544 $\pm$ 0.057 & 0.1 $\pm$ 0.0 & 62.5 $\pm$ 2.2 & 7.5 $\pm$ 0.9 & 19.9 $\pm$ 4.9 & 80.0 $\pm$ 4.9 & 42.5 $\pm$ 0.3 \\
Reference (always OK) & 11.1 $\pm$ 0.0 & 0.500 $\pm$ 0.000 & 0.0 $\pm$ 0.0 & 0.0 $\pm$ 0.0 & 100.0 $\pm$ 0.0 & 100.0 $\pm$ 0.0 & 0.0 $\pm$ 0.0 & 50.0 $\pm$ 0.0 \\
        \bottomrule
    \end{tabular*}
    \par\vspace{3pt}{\scriptsize\renewcommand{\arraystretch}{1}\raggedright Values are in percent, except for the AUROC, and are given as mean $\pm$ standard deviation over three seeds. Unknown metrics are first averaged over the held-out failure modes. Unknown detection is at a 5\% false positive rate. The sorting loss is the mean of the four sorting errors. RF + PCA flag is random forests with the PCA unknown flag; it has no classifier or score of its own, so it only has sorting errors. Bold marks the best method per column, excluding the reference.\par}
\end{table*}

\subsection{Latent Analysis}
\label{subsec:latent}
\begin{figure*}[!t]
    \centering
    \includegraphics[width=\textwidth]
        {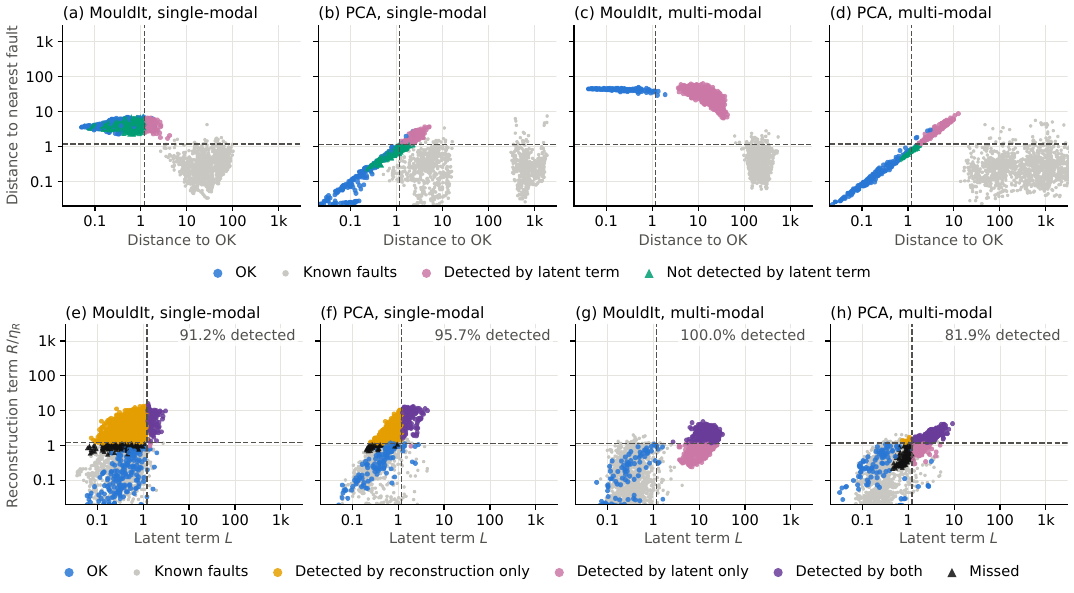}
    \caption{Novelty score of MouldIt and PCA with secondary pressure rise held out as the unknown fault, at 16000 cycles and seed 42. Distances are measured in the 32-dimensional latent space of MouldIt and on the 32 components of PCA. Panels (a)--(d) show the distance of each test cycle to the OK class and to its nearest known fault, each in units of that class's 95\% boundary. Panels (e)--(h) show the two terms of the novelty score, the latent term $L$ and the reconstruction term $R/\eta_R$. Dashed lines mark the threshold $\Theta$ in all panels: the latent term detects a cycle only past both lines in (a)--(d), and a cycle is missed only in the lower left box in (e)--(h). Unknown cycles are colored by whether the latent term detects them in (a)--(d), and by whether the reconstruction term only, the latent term only, or both detect them in (e)--(h). The percentages give the share of unknown cycles flagged in this fold. We draw about half of the cycles for each label in the legend.}
    \label{fig:latent}
\end{figure*}

\begin{table}[!t]
    \centering
    \caption{Detection of Unknown Faults at 16000 Cycles}
    \label{tab:latent}
    \renewcommand{\arraystretch}{1.0}
    \scriptsize
    \setlength{\tabcolsep}{2.5pt}
    \begin{tabular*}{\columnwidth}{@{\extracolsep{\fill}}lcccccccc@{}}
        \toprule
        & \multicolumn{2}{c}{Detected} & \multicolumn{2}{c}{Latent only} & \multicolumn{2}{c}{Recon. only} & \multicolumn{2}{c}{Both} \\
        \cmidrule(lr){2-3} \cmidrule(lr){4-5} \cmidrule(lr){6-7} \cmidrule(l){8-9}
        Held-out mode & M & P & M & P & M & P & M & P \\
        \midrule
\multicolumn{9}{@{}l}{\textit{Single-modal}}\\
Machine drift & 42 & 39 & 28 & 19 & 10 & 3 & 4 & 17 \\
Secondary pressure rise & 92 & 95 & 0 & 0 & 64 & 68 & 27 & 27 \\
Underpacking/overpacking & 69 & 33 & 29 & 33 & 1 & 0 & 40 & 0 \\
Time shift & 100 & 100 & 0 & 0 & 3 & 0 & 97 & 100 \\
Fast gate freeze & 100 & 100 & 0 & 0 & 11 & 0 & 89 & 100 \\
\midrule
\multicolumn{9}{@{}l}{\textit{Multi-modal}}\\
Secondary pressure rise & 100 & 83 & 28 & 24 & 0 & 1 & 72 & 58 \\
Underpacking/overpacking & 100 & 100 & 6 & 100 & 0 & 0 & 94 & 0 \\
Machine drift & 100 & 100 & 8 & 83 & 0 & 0 & 92 & 17 \\
Temperature dynamics & 100 & 99 & 1 & 18 & 0 & 0 & 99 & 81 \\
Inter-sensor inconsistency & 100 & 100 & 0 & 0 & 0 & 0 & 100 & 100 \\
Fast gate freeze & 100 & 100 & 0 & 0 & 0 & 0 & 100 & 100 \\
Distorted US response & 100 & 100 & 0 & 0 & 0 & 0 & 100 & 100 \\
Time shift & 100 & 100 & 0 & 0 & 0 & 0 & 100 & 100 \\
        \bottomrule
    \end{tabular*}
    \par\vspace{3pt}{\scriptsize\renewcommand{\arraystretch}{1}\raggedright Mean percentage of each held-out mode's cycles over three seeds, hardest first. M: MouldIt, P: PCA. Each detected cycle is flagged by the latent term of the novelty score only, its reconstruction term only, or both, so the three sum to Detected, up to rounding.\par}
\end{table}

To understand these results, we split the novelty score~\eqref{eq:novelty} of every fold into its latent and reconstruction terms, for MouldIt and PCA (Table~\ref{tab:latent}, Fig.~\ref{fig:latent}).

\subsubsection{OK and Known Faults}
At 16000 cycles, only 1.0\% (single-modal) and 0\% (multi-modal) of the OK test cycles lie inside the 95\% boundary of a known fault in the MouldIt latent space (Fig.~\ref{fig:latent}(a), (c)). In the PCA space, 96.0\% and 97.6\% do (Fig.~\ref{fig:latent}(b), (d)). In units of these boundaries, the median distance of an OK cycle to its nearest known fault is 10.1 and 25.3 in MouldIt, but only 0.17 and 0.15 in PCA. PCA components follow the largest variance in the data, whether or not it separates the classes, while MouldIt is also trained to classify from its latent space. Since PCA predicts the nearest class (Section~\ref{subsec:novelty}), this explains why it rejects most good parts (Figs.~\ref{fig:sorting_pressure_only}(a) and~\ref{fig:sorting_multi-modal}(a)). The separation needs data: at 200 single-modal cycles, 60.6\% of the OK cycles still lie inside the boundary of a known fault in MouldIt.

\subsubsection{Unknown Faults}
With more data, the MouldIt latent space places unknown faults outside the known classes. The share of unknown cycles inside the boundary of a known class falls from 88.3\% at 200 cycles to 33.1\% at 16000 cycles with pressure alone, and from 14.4\% to 0\% with all channels. With less data, MouldIt therefore relies more on reconstruction error: at 1000 single-modal cycles, 54.7\% of the unknown cycles are flagged by the reconstruction term alone, and at 16000 only 17.8\%, as most are then flagged by both terms.

\subsubsection{Single-Modal Limits}
With pressure alone, underpacking/overpacking, secondary pressure rise, and machine drift stay below full detection (Table~\ref{tab:latent}). Underpacking/overpacking scales the pressure (Section~\ref{methods:single}), a change that lies almost entirely in the linear span of the PCA components. PCA therefore reconstructs this unknown fault even better than the known cycles: its median reconstruction term is 0.007, and detecting the fault from this term alone gives an unknown AUROC of 0.410. For MouldIt, the same values are 1.01 and 0.806, which makes this failure mode the main source of its single-modal lead in unknown detection. MouldIt places 66.8 $\pm$ 26.8\% of the secondary pressure rise cycles inside the OK boundary (Fig.~\ref{fig:latent}(a)). These cycles are caught almost only by the reconstruction term, and those below its threshold are missed (Fig.~\ref{fig:latent}(e)). Machine drift stays close to the known classes, and neither term separates it well for either method.

\subsubsection{Multi-Modal Gain}
With all channels, MouldIt fully detects underpacking/overpacking, secondary pressure rise, and machine drift. For machine drift, the added channels carry direct evidence of the fault, since our generator raises the temperature baseline and adds noise there (Section~\ref{methods:multi}). Secondary pressure rise and underpacking/overpacking, however, leave the added channels unchanged (Fig.~\ref{fig:synthetic_multi-modal}(c), (e)). Yet the latent space now moves secondary pressure rise away from OK (Fig.~\ref{fig:latent}(c), (g)). The gain on these two modes therefore cannot come from direct evidence of the fault in the added channels. The richer set of known classes or the coupling between channels may shape the latent space, but we have not tested either.

\subsection{Reconstruction Fidelity}
\begin{table}[!t]
    \centering
    \caption{Reconstruction Fidelity at 16000 Cycles}
    \label{tab:fidelity}
    \renewcommand{\arraystretch}{1.0}
    \scriptsize
    \setlength{\tabcolsep}{1pt}
    \begin{tabular*}{\columnwidth}{@{\extracolsep{\fill}}lcccc@{}}
        \toprule
        & \multicolumn{2}{c}{Single-modal} & \multicolumn{2}{c}{Multi-modal} \\
        \cmidrule(lr){2-3} \cmidrule(l){4-5}
        Metric & PCA & MouldIt & PCA & MouldIt \\
        \midrule
\multicolumn{5}{@{}l}{\textit{Known faults}}\\
Deviation $R^2$ & \textbf{0.997 $\pm$ 0.000} & 0.994 $\pm$ 0.001 & \textbf{0.995 $\pm$ 0.000} & 0.994 $\pm$ 0.001 \\
DTW shape fidelity & \textbf{0.984 $\pm$ 0.000} & 0.976 $\pm$ 0.003 & \textbf{0.989 $\pm$ 0.000} & 0.986 $\pm$ 0.001 \\
TDI & \textbf{0.016 $\pm$ 0.000} & 0.049 $\pm$ 0.005 & \textbf{0.001 $\pm$ 0.000} & 0.003 $\pm$ 0.001 \\
\midrule
\multicolumn{5}{@{}l}{\textit{Unknown fault}}\\
Deviation $R^2$ & \textbf{0.850 $\pm$ 0.001} & 0.485 $\pm$ 0.180 & \textbf{0.779 $\pm$ 0.001} & 0.433 $\pm$ 0.043 \\
DTW shape fidelity & 0.22 $\pm$ 0.06 & \textbf{0.26 $\pm$ 0.23} & \textbf{-0.20 $\pm$ 0.02} & -2.58 $\pm$ 1.44 \\
TDI & \textbf{0.048 $\pm$ 0.001} & 0.083 $\pm$ 0.003 & \textbf{0.019 $\pm$ 0.000} & 0.033 $\pm$ 0.002 \\
        \bottomrule
    \end{tabular*}
    \par\vspace{3pt}{\scriptsize\renewcommand{\arraystretch}{1}\raggedright Mean $\pm$ standard deviation over three seeds, on faulty test cycles; for the unknown fault, averaged over the held-out failure modes. Deviation $R^2$: share of the deviation from the reference cycle that the reconstruction recovers. DTW shape fidelity: the same under dynamic time warping. TDI: temporal distortion index. Bold: better method per setting. Higher is better, except for the TDI.\par}
\end{table}

We measure how well MouldIt and PCA reconstruct the deviation that a fault causes (Table~\ref{tab:fidelity}). The deviation $R^2 = 1 - \sum(\hat{x} - x)^2 / \sum(x - x_{\mathrm{ref}})^2$ is the share of this deviation from the reference cycle that the reconstruction recovers, where 0 means no better than copying the reference. The DTW shape fidelity is the same measure under dynamic time warping (DTW), following the shape and timing split of \cite{leguen2019shape}, and the temporal distortion index (TDI) \cite{friasparedes2017assessing} measures the remaining timing error, where 0 means none. The averages of the unknown DTW shape fidelity are dominated by the time shift mode, which is negative for both methods.

PCA is clearly ahead on unknown faults. Our data is close to linear around one reference cycle, PCA directly minimizes the squared error, and the MouldIt latent space also serves classification. High fidelity on unknown faults is not a goal in itself, since their poorer reconstruction is what makes the MouldIt reconstruction term useful for unknown detection.

\subsection{Limitations}
All our cycles derive from one real cycle of one part and one material, and the magnitudes of our failure modes are modeling choices rather than calibrated effects (Section~\ref{subsec:synthetic-data-generation}). Our results therefore show what MouldIt can learn from process-informed synthetic data, not how it performs on real failures.

The sorting loss (Section~\ref{sec:experiments}) weighs its four rates equally, and our classes are balanced, whereas real production has few faults and unequal costs. Our operating point of 5\% FPR is high for a production line.

At small sizes, our results are noisy. At 50 cycles, a class has at most five training cycles and one or two validation cycles, and the novelty thresholds are set from these few cycles. We selected the KLD and classification weights on one seed at a size of 1000, on folds that are also part of our final sweep, which is a mild selection on test data. Our novelty score (Section~\ref{subsec:novelty}) fits one Gaussian per class, although some classes are bimodal by design, for example underpacking and overpacking. Such a Gaussian also covers the gap between the two modes, so an unknown fault landing there looks familiar. Finally, we chose the network size and the latent dimension empirically. We also did not record a formal ablation, so we cannot yet attribute the latent separation to the BCE rather than to the MSE and KLD, which train the same encoder.

Future work should run MouldIt on real cycles with real failures, and could use its learned representations for process control. Future work should also train it without the BCE term, to test whether the BCE causes the latent separation, and with four channels on only the single-modal failure modes, to test whether the multi-modal gain comes from the richer set of known classes or from the coupling between channels.

\section{Conclusion}
We introduced MouldIt, a multi-objective temporal VAE for in-mould signals, and studied how its performance depends on the amount of training data and the input modalities, using synthetic datasets generated from a single real cycle. A learned representation outperforms the classical methods once there is enough data or more than one modality, and a single model can then classify parts, detect unknown faults, and sort parts. At 16000 cycles, the part sorting loss of MouldIt is the lowest of all compared methods, at 6.0 $\pm$ 0.7\% with pressure alone and 1.0 $\pm$ 0.3\% with all four channels; in the latter setting, random forests with the PCA unknown flag are on par. Whether this holds in real production remains the main open question.

\section*{Acknowledgment}
This work was carried out in part within the research project FORinFPRO (Project No. AZ-1596-23), funded by the Bayerische Transformations- und Forschungsstiftung. The author thanks R. Kini and L. Schweizer for the dataset used in this work.
The author conceptualized this work, and generative AI was not used to formulate its research questions. Claude, Perplexity, and ChatGPT were used for literature search and brainstorming. Under the author's direction, Claude was also used to write code, following the author's code patterns, to prepare the figures, to co-create the Discussion, Conclusion, figure captions, and novelty score description from the author's results, insights, post-results abstract, and instructions, and to edit the wording and formatting of the remaining text. The author reviewed and revised all AI-assisted work over many iterations and takes responsibility for the content.

\bibliographystyle{IEEEtran}
\bibliography{mybibfile}

\end{document}